\documentclass[letterpaper, 10 pt, conference]{ieeeconf}  

\IEEEoverridecommandlockouts                              
\usepackage{graphicx} 
\usepackage{float}
\usepackage{subfigure}

\usepackage[noadjust]{cite}

\usepackage{amsmath} 
\usepackage{color}
\usepackage{url}

\title{\LARGE \bf
AU-AIR: A Multi-modal Unmanned Aerial Vehicle Dataset for Low Altitude Traffic Surveillance}

\author{Ilker Bozcan, Erdal Kayacan
\thanks{I. Bozcan and E. Kayacan are with the Department of Engineering, Aarhus University,
        8000 Aarhus C, Denmark
        {\tt\small \{ilker, erdal\} at eng.au.dk}}%
}

\begin{document}

\maketitle
\thispagestyle{empty}
\pagestyle{empty}

\begin{abstract}

Unmanned aerial vehicles (UAVs) with mounted cameras have the advantage of capturing aerial (bird-view) images. The availability of aerial visual data and the recent advances in object detection algorithms led the computer vision community to focus on object detection tasks on aerial images. As a result of this, several aerial datasets have been introduced, including visual data with object annotations. UAVs are used solely as flying-cameras in these datasets, discarding different data types regarding the flight (e.g., time, location, internal sensors). In this work, we propose a multi-purpose aerial dataset (AU-AIR) that has multi-modal sensor data (i.e., visual, time, location, altitude, IMU, velocity) collected in real-world outdoor environments. The AU-AIR dataset includes meta-data for extracted frames (i.e., bounding box annotations for traffic-related object category) from recorded RGB videos. Moreover, we emphasize the differences between natural and aerial images in the context of object detection task. For this end, we train and test mobile object detectors (including YOLOv3-Tiny and MobileNetv2-SSDLite) on the AU-AIR dataset, which are applicable for real-time object detection using on-board computers with UAVs. Since our dataset has diversity in recorded data types, it contributes to filling the gap between computer vision and robotics. The dataset is available at \url{https://bozcani.github.io/auairdataset}.

\end{abstract}

\section{INTRODUCTION}

Unmanned aerial vehicles (UAVs) are extensively used as flying platforms of sensors for different domains such as traffic surveillance \cite{puri2005survey}, managing the urban environment \cite{gallacher2016drones}, package delivery \cite{mehndiratta2019constrained} or aerial cinematography \cite{doi:10.1002/rob.21931}. For these applications, UAVs are equipped with mounted cameras and mainly gather visual data of the environment. Then, computer vision algorithms are applied to aerial visual data to extract high-level information regarding the environment.

Object detection is one of the most studied problems in computer vision. The recent advances in deep learning (variants of convolutional neural networks (CNNs) mainly) have led to breakthrough object detection performances with the availability of large datasets and computing power. Since these methods require a large number of training samples, several datasets (e.g., COCO \cite{lin2014microsoft}, Pascal VOC \cite{everingham2010pascal}) have been introduced for benchmarking for the object detection task. The samples in these datasets consist of natural images that are mainly captured by handheld cameras.  The significant differences between natural and aerial images (such as object layouts and sizes) cause these object detectors to have trouble to find objects in aerial images. Therefore, several datasets (e.g., \cite{zhuvisdrone2018, du2018unmanned, hsieh2017drone, robicquet2016learning, mueller2016benchmark, collins2005open, highDdataset}) have been introduced in recent years as a benchmark for object detection in aerial images.

\begin{figure}[!hbt]
    \centering
    \includegraphics[width=0.49\textwidth]{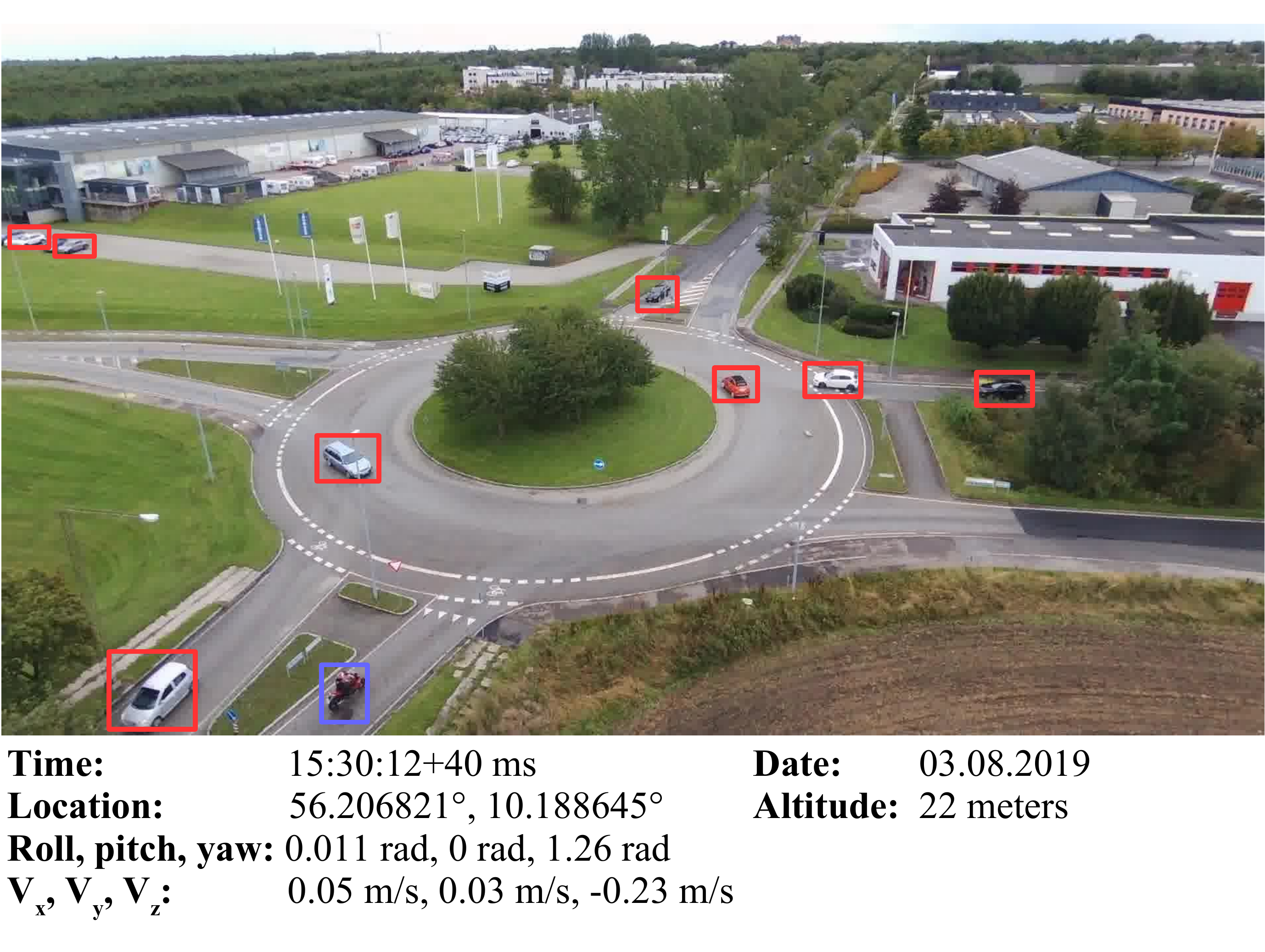}
    \caption{In the AU-AIR dataset, extracted frames are labeled with object annotations, time stamp, current location, altitude, and velocity of the UAV, and the rotation data read from the IMU sensor. [Best viewed in color]}
    \label{fig:intro}
\end{figure}

Besides visual data gathered by a camera, the data from other sensors might give crucial information about the environment. The use of UAVs as only flying cameras cut off the potential advance in multi-modal object detection algorithms for aerial applications. For instance, the recent advances in perception for autonomous driving have brought new datasets such as \cite{geiger2013vision, caesar2019nuscenes, choi2018kaist} including multi-modal data (e.g., RGB images, Global Positioning System (GPS) coordinates, inertial measurement unit (IMU) data). Although the data fusion for object detection is still open research topic \cite{feng2019deep}, these multi-modal datasets allow a benchmark for further research. However, to the best of our knowledge, there is no such multi-modal dataset collected in a real-world outdoor environment for UAVs.

In this work, we present a multi-modal UAV dataset (The AU-AIR dataset) in order to push forward the development of computer vision and robotic algorithms targeted at autonomous aerial surveillance. The AU-AIR dataset meets vision and robotics for UAVs having the multi-modal data from different on-board sensors. The dataset consists of 8 video streams (over 2 hours in total) for traffic surveillance. The videos mainly are recorded at Skejby Nordlandsvej and P.O Pedersensvej roads (Aarhus, Denmark). The dataset includes aerial videos, time, GPS coordinates and the altitude of the UAV, IMU data, and the velocity. The videos are recorded at different flight altitudes from 5 meters to 30 meters and in different camera angles from 45 degrees to 90 degrees (i.e., complete bird-view images that the camera is perpendicular to the Earth). Instances belonging to different object categories related to the traffic surveillance context are annotated with bounding boxes in video frames. Moreover, each extracted video frame is labeled with the flight data (See Fig. \ref{fig:intro}).

The whole dataset includes 32,823 labeled video frames with object annotations and the corresponding flight data. Eight object categories are annotated including person, car, van, truck, motorbike, bike, bus, trailer. The total number of annotated instances is 132,034. The dataset is split into 30,000 training-validation samples and 2,823 test samples. 

In this work, we emphasize differences between aerial and natural images in the context of object detection tasks. To this end, we compare image samples and object instances between the AU-AIR dataset and the COCO dataset \cite{lin2014microsoft}. In our experiments, we train and evaluate two mobile object detectors (including YOLOv3-tiny \cite{redmon2018yolov3} and MobileNetv2-SSD Lite \cite{sandler2018mobilenetv2} on the AU-AIR dataset. We form a baseline, including mobile object detectors since we focus on real-time performance and the applicability of object detection task onboard computers mounted on UAV.

\subsection{Related Work}

In recent years, several drone datasets have been introduced for object detection tasks (\cite{zhuvisdrone2018, du2018unmanned, hsieh2017drone, robicquet2016learning, mueller2016benchmark, collins2005open, highDdataset}). Zhu et al. \cite{zhuvisdrone2018} propose a UAV dataset (VisDrone) consisting of visual data and object annotations in images and frames. In the VisDrone dataset, object instances belonging the certain categories are annotated by bounding boxes and category labels. Besides object annotations, VisDrone includes some vision-related attributes such as the visibility of a scene, occlusion status. Du et al. \cite{du2018unmanned} propose a benchmark dataset for object detection and tracking in aerial images. The dataset also includes meta information regarding the flight altitude. Hsieh et al. \cite{hsieh2017drone} propose a UAV-based counting dataset (CARPK) including object instances that belong to the car category. Robicquet et al. \cite{robicquet2016learning} introduce a UAV dataset (Stanford) that collects images and videos of six types of objects in the Stanford campus area. In this dataset, some of the object categories dominate the dataset having a high number of samples, whereas the remaining object categories have significantly less number of instances. Mueller et al. \cite{mueller2016benchmark} propose synthetic dataset created by a simulator for target tracking with a UAV. Collins et al. \cite{collins2005open} introduce a benchmarking website (VIVID) with an evaluation dataset collected under the DARPA VIVID program. Krajewski et al. propose an aerial dataset collected from highways, including object bounding boxes and labels of vehicles.

These datasets are annotated by common objects in an environment such as humans and different types of vehicles (e.g., car, bike, van). However, they only include visual data and bounding box annotations for objects and discard other sensory data. Among these studies, only UAVDT \cite{du2018unmanned} includes an attribute that gives limited information about the flight altitude (i.e., labels such as "low-level", "mid-level" and "high-level").

Fonder et al. \cite{fonder2019mid} propose a synthetic dataset (Mid-Air) for low altitude drone flights in unstructured environments (e.g., forest, country). It includes multi-modal data regarding the flight (e.g., visual, GPS, IMU data) without any annotations for visual data.

There are also multi-modal drone datasets in the literature (\cite{fonder2019mid, antonini2018blackbird, burri2016euroc, majdik2017zurich, sun2018robust}). However, the visual data are not collected for object detection since the main focus of these studies is the UAV navigation. Therefore, these datasets do not have object annotations. The comparison of existing datasets is given in Table \ref{tbl:comparison}.

\subsection{Contribution}
Looking also at the summary of the existing studies in Table \ref{tbl:comparison}, the followings are the main contributions of this work:

\begin{itemize}

\item To the best of our knowledge, the AU-AIR dataset is the first multi-modal UAV dataset for object detection. The dataset includes flight data (i.e., time, GPS, altitude, IMU data) in addition to visual data and objects annotations.

\item Considering the real-time applicability, we form a baseline training and testing mobile object detectors with the AU-AIR dataset. We emphasize the differences between object detection in aerial images and natural images.

\end{itemize}


\begin{table*}[hbt!]
\caption{Comparison with existing UAV datasets. 
\label{tbl:comparison}}
\begin{center}
\footnotesize
\begin{tabular}{c|ccccccccc}\hline
Dataset & Environment & Data type & Visual data & Object annotations & Time & GPS & Altitude  & Velocity & IMU data \\  \hline \hline

VisDrone \cite{zhuvisdrone2018} & outdoor & real & yes &  yes & no & no & no & no & no  \\
UAVDT \cite{du2018unmanned} & outdoor & real & yes & yes & no & no & partial & no & no \\
CARPK \cite{hsieh2017drone} & outdoor & real & yes & yes & no & no & no & no & no \\
Stanford \cite{robicquet2016learning} & outdoor & real & yes & yes & no & no & no & no & no \\
UAV123 \cite{mueller2016benchmark} & outdoor & synthetic & yes & yes & no & no & no & no & no \\
VIVID \cite{collins2005open} & outdoor & real & yes & yes & no & no & no & no & no \\
highD \cite{highDdataset} & outdoor & real & yes & yes & no & no & no & no & no \\
Mid-Air \cite{fonder2019mid} & outdoor & synthetic & yes & no & yes & yes & yes & yes & yes \\
Blackbird \cite{antonini2018blackbird} & indoor & real & yes & no & yes & yes & yes & yes & yes \\
EuRoC MAV \cite{burri2016euroc} & indoor & real & yes & no & yes & yes & yes & yes & yes \\
Zurich Urban MAV \cite{majdik2017zurich} & outdoor & real & yes & no & yes & yes & yes & yes & yes \\
UPenn Fast Flight \cite{sun2018robust} & outdoor & real & yes & no & yes & yes & yes & yes & yes \\

\hline
\hline
\textbf{AU-AIR}  & \textbf{outdoor} & \textbf{real} & \textbf{yes} & \textbf{yes} & \textbf{yes} & \textbf{yes} & \textbf{yes} & \textbf{yes} & \textbf{yes} \\

\hline

\end{tabular}
\end{center}

\end{table*}

\section{OBJECT DETECTION IN NATURAL IMAGES VS AERIAL IMAGES}
The availability of large amounts of data and processing power enables deep neural networks to achieve state-of-the-art results for object detection. Currently, deep learning-based object detectors are separated into two groups. The first group consists of region-based CNNs that ascend on image classifiers. Region-based CNNs propose image regions that are likely to contain an object and classify the region into a predefined object category. The second group has only one stage converting to the object detection problem into the bounding box prediction for objects, without re-purposing image classifiers. Faster-R-CNN \cite{ren2015faster} is one of the well-known models belonging to the first group, YOLO \cite{redmon2016you} and SSD \cite{liu2016ssd} are the popular object detectors that belong to the second group.

Deep learning-based object detectors have trained and performed on large datasets such as COCO \cite{lin2014microsoft} and PASCAL \cite{everingham2010pascal}. These datasets include natural images that contain a single object or multi objects in their natural environments. Most of the images in these datasets are captured by humans using a handheld camera so that the vast majority of images have side-view. There are challenges of the object detection in natural images such as occlusion, illumination changes, rotation, low resolution,  crowd existence of instances.

\begin{figure}[!t]
    \centering
    \includegraphics[width=0.49\textwidth]{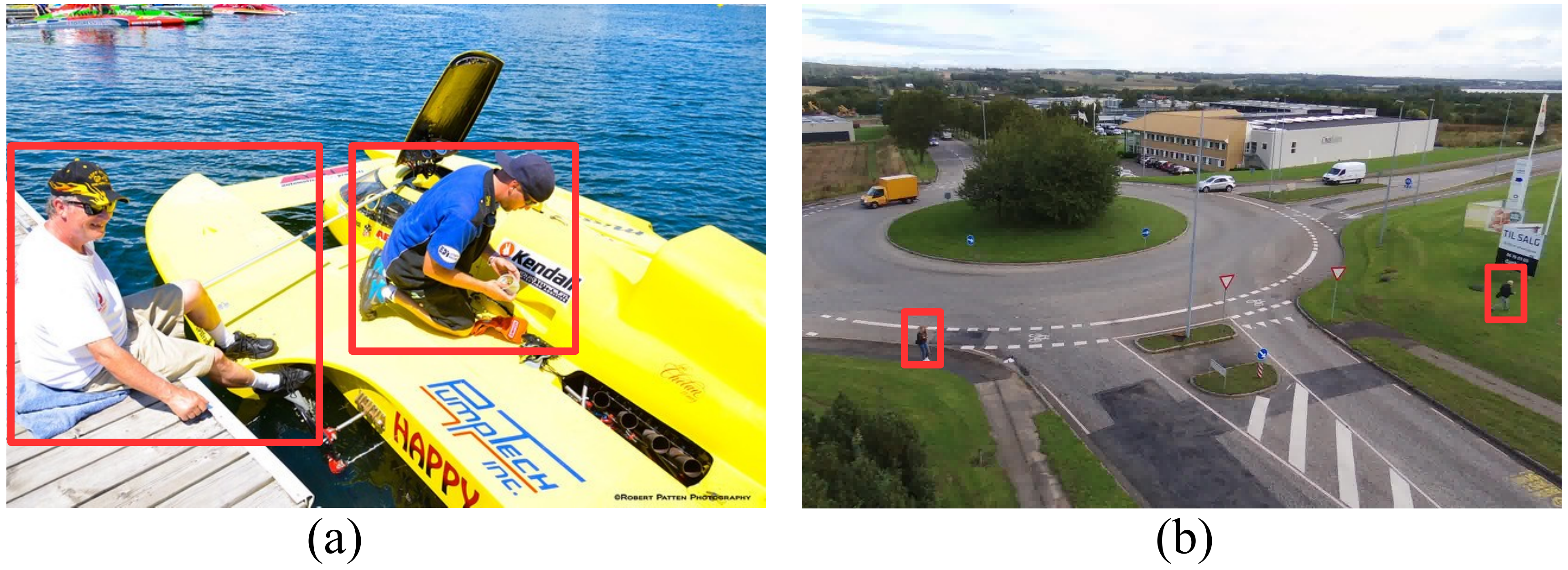}
    \caption{Sample images, including human category from COCO (a) and AU-AIR (b) datasets. Human instances occupy a significantly larger area in the natural image (a) than in the aerial image (b). Figure (b) is captured at an altitude of 12 meters that is close to the lowest value of flight altitude range (5 meters). Objects tend to occupy a much smaller area when the altitude increases. [Best viewed in color]}
    \label{fig:human_compare}
\end{figure}

Aerial images have different characteristics from natural images due to having a bird's-eye view. First of all, objects in natural images are much larger than their counterparts in aerial images. For example, an object category such as humans may occupy a large number of pixels in natural images. However, it may have a few numbers of pixels in an aerial image that is quite challenging to detect for object detectors (See Fig. \ref{fig:human_compare}). Moreover, aerial images can be fed to a network with higher dimensions that increases computational cost in order to prevent the diminishing of pixels belonging to small objects.

\begin{figure}[!t]
    \centering
    \includegraphics[width=0.49\textwidth]{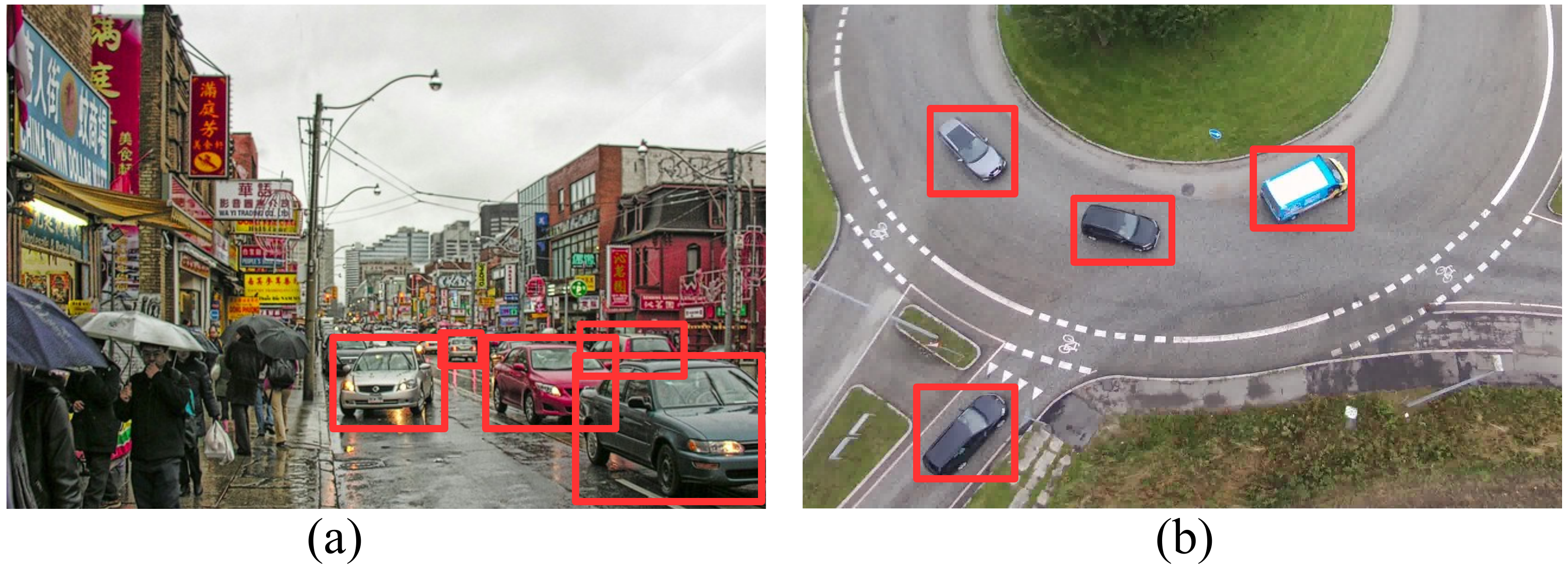}
    \caption{Sample images, including car category from COCO (a) and AU-AIR (b) datasets. Car instances are occluded by other cars in a natural image (a) that is captured by the handheld camera, but they are less likely to occur due to other instances in an aerial image (b). [Best viewed in color]}
    \label{fig:occlusion_compare}
\end{figure}

Secondly, an occlusion is observed in different conditions for natural and aerial images. In natural images, an object instance may be occluded by another foreground object instance (e.g., a human in front of a car). However, objects in aerial images are less likely to be occluded by other foreground objects (especially, bird-view images captured by a camera that is perpendicular to the Earth). (See Fig. \ref{fig:occlusion_compare}. 

\begin{figure}[!t]
    \centering
    \includegraphics[width=0.49\textwidth]{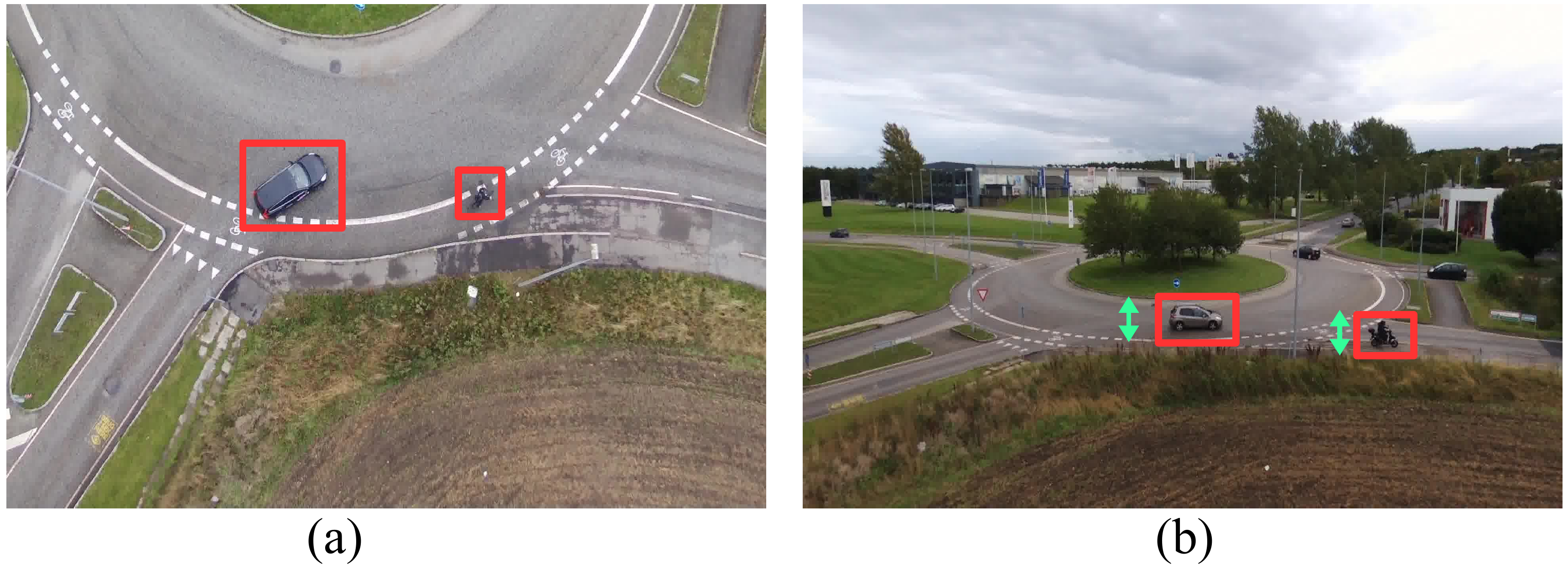}
    \caption{Sample images from the AU-AIR dataset. The object height information is lost in images captured with a complete bird-view angle (a), whereas it can be inferred in images with a side-view angle (b). The heights of the bounding box in (b) give a clue about objects' height (green arrows). [Best viewed in color]}
    \label{fig:perspective_compare}
\end{figure}

Thirdly, the perspective in aerial images makes appearances of objects short and squat. This fact diminishes the information regarding an object height (See Fig. \ref{fig:perspective_compare}). Moreover, although aerial images can supply more contextual information about an environment by a broader view angle, the object instances may be amid cluttered.

Lastly, having a drone to capture aerial images, the altitude changes during the flight can cause varieties in object size and appearance in aerial images. Therefore, a recording of aerial videos at different altitudes may change the levels of challenges mentioned above. 




\section{AU-AIR -- THE MULTI-MODAL UAV DATASET}


To address the challenges mentioned in Section II, we propose a multi-modal drone dataset (AU-AIR) including videos, object annotations in the extracted frames and sensor data for the corresponding frames. The data are captured by low-level flight (max. 30 meters) and for the scenario of a traffic surveillance. The  AU-AIR dataset consists of video clips, sensor data, and object bounding box annotations for video frames.

\subsection{UAV Platform}
We have used a quadrotor (Parrot Bebop 2) to capture the videos and record the flight data. An on-board camera has recorded the videos with a resolution of $1920 \times 1080$ pixels at 30 frames per second (fps). The sensor data have been recorded for every 20 milliseconds.

\subsection{Dataset Collection}
The AU-AIR dataset consists of 8 video clips (approximately in 2 hours of a total length) with 32,823 extracted frames. All videos are recorded for a scenario of aerial traffic surveillance at the intersection of Skejby Nordlandsvej and P.O Pedersensvej (Aarhus, Denmark) on windless days. Moreover, the videos cover various lighting conditions due to the time of the day and the weather conditions (e.g., sunny, partly sunny, cloudy).

Capturing an aerial video with a UAV brings different challenges for visual surveillance that are significantly different from natural images. To add these challenges in our dataset, we have captured the videos in different flight altitudes and camera angles. The flight altitude changes between 10 meters to 30 meters in the videos and the camera angle is adjusted from 45 degrees to 90 degrees (perpendicular to the Earth). An increase in the camera angle makes object detection task more challenging since images get differ from natural images.

Although the videos have been recorded with 30 fps, we have extracted five frames for every second in order to prevent the redundant occurrence of frames. Both of raw videos and extracted frames have a resolution of $1920\times1080$ pixels.

\subsection{Visual Data and Annotation}
Considering a traffic surveillance scenario, we have manually annotated specific object categories in the frames. For annotation, we used a bounding box and object category index for each instance. The annotated object categories include eight types of objects which highly occur during the traffic surveillance: person, car, bus, van, truck, bike, motorbike, and trailer.

\begin{figure}[!hbt]
    \centering
    \includegraphics[width=0.45\textwidth]{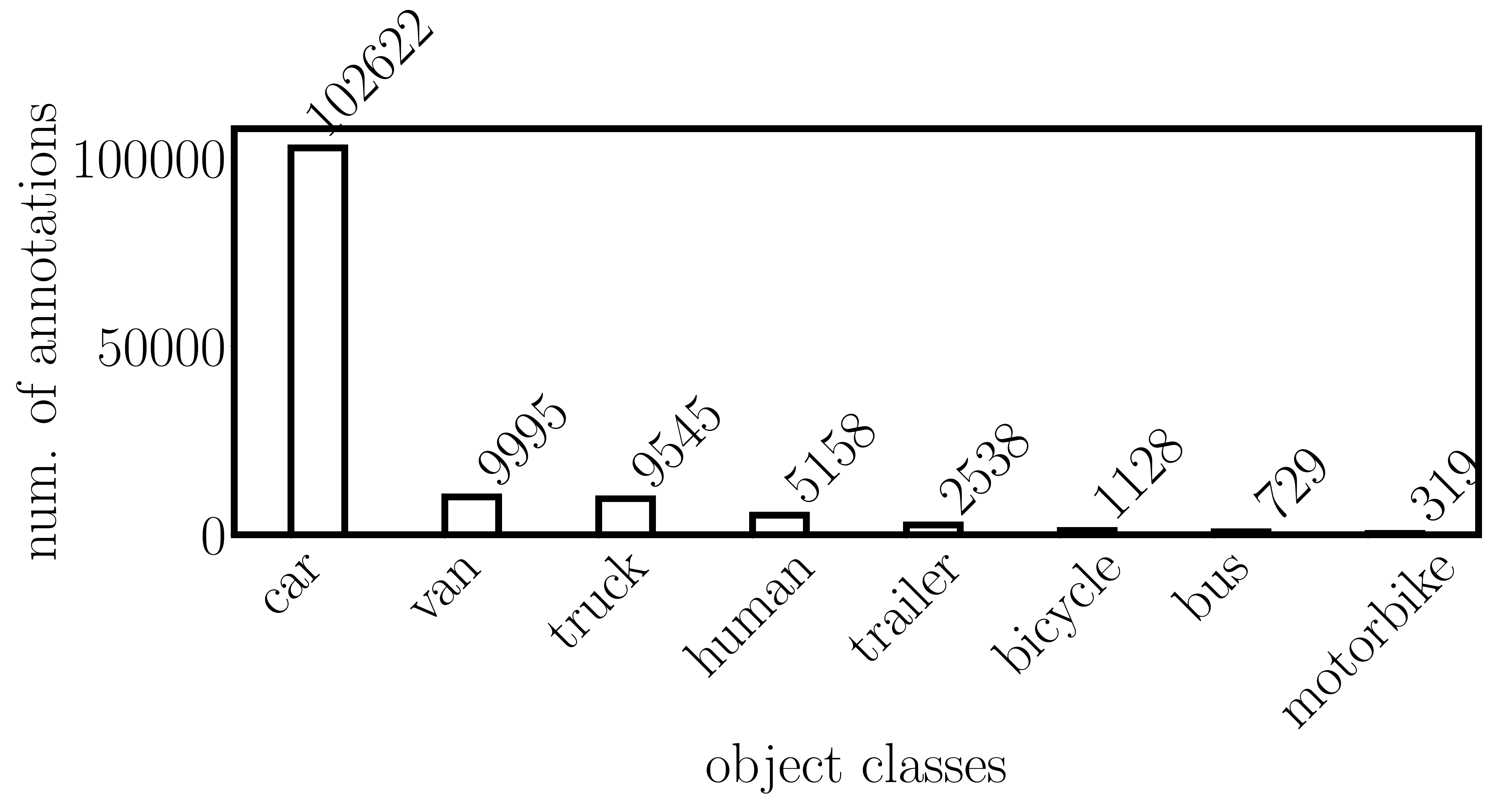}
    \caption{Distribution of annotations across object classes.}
    \label{fig:classes_vs_numsamples}
\end{figure}

For annotation, we employed workers on Amazon’s Mechanical Turk (AMT) \cite{turk2012amazon}. In order to increase the labeling quality, three workers annotated the same frame separately. Then, we combined annotations if they have the same object labels, and whose bounding boxes overlap more than a certain threshold. We chose a threshold as a value of 0.75 experimentally. In case this condition is not satisfied, we manually fine-tuned the bounding boxes and class labels.
The category distribution over the dataset can be seen in Fig. \ref{fig:classes_vs_numsamples}. In the context of traffic surveillance, cars appear significantly more than other classes, and three vehicle types (car, van, truck) have a major portion of annotated bounding boxes.

\begin{figure}[t!]
    \centering
    \includegraphics[width=0.42\textwidth]{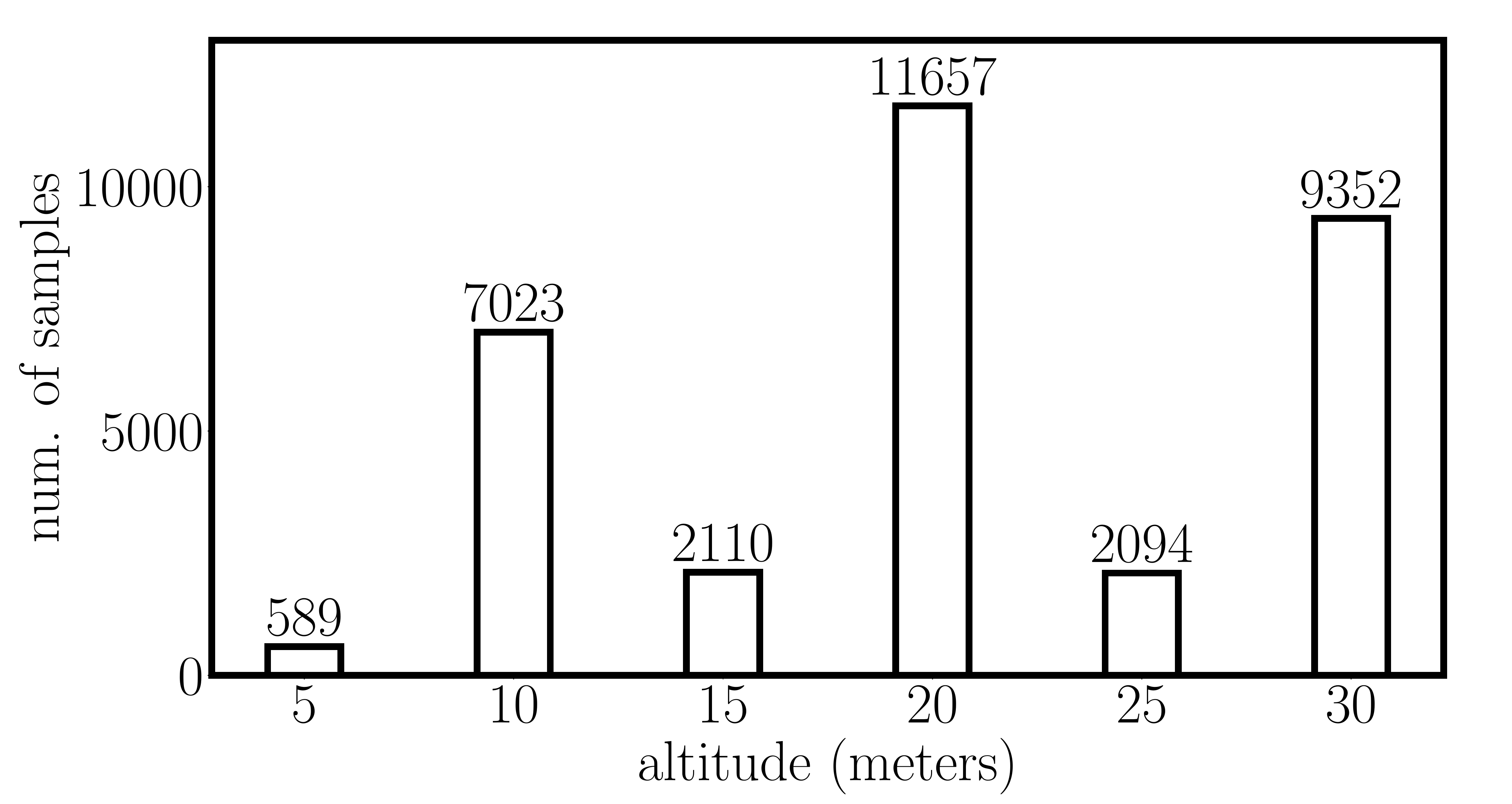}
    \caption{Distribution of samples across flight altitudes.}
    \label{fig:altitude_vs_numsamples}
\end{figure}

The AU-AIR dataset includes frames that are captured in different flight altitudes (See Fig. \ref{fig:altitude_vs_numsamples}). We recorded the data mainly for 10 meters, 20 meters, and 30 meters with different camera angles from 45 degrees to 90 degrees.

\subsection{Sensor Data}
In addition to visual data and object annotations, the AU-AIR dataset includes sensor data that are logged during the video recording. In the dataset, we have the following attributes for each extracted frame:
\begin{itemize}
    \item $d$: current date of a frame
    \item $t$: current time stamp of the frame
    \item $la$: latitude of the UAV (read from GPS sensor)
    \item $lo$: longitude of the UAV (read from GPS sensor)
    \item $a$: altitude of the UAV (read from altimeter)
    \item $\phi$: UAV roll angle (rotation around the x axis) (read from IMU sensor)
    \item $\theta$: UAV pitch angle (rotation around the y axis) (read from IMU sensor)
    \item $\psi$: UAV yaw angle (rotation around the z axis) (read from IMU sensor)
    \item $V_x$: speed on the x axis
    \item $V_y$: speed on the y axis
    \item $V_z$: speed on the z axis
\end{itemize}

Table \ref{tbl:sensor_data} shows unit values and ranges for each attribute except the date. The date ($d$) has a format of MMDDYYYY-HHMMSS where MM, DD, YYYY, HH, MM, SS indicates the month, day, year, hour, minutes, and second, respectively.

\begin{table}[ht]
\caption{Sensor data types in the dataset. 
\label{tbl:sensor_data}}
\begin{center}
\footnotesize
\begin{tabular}{cccc}\hline
  Data & Unit & Min. value & Max. value  \\  \hline \hline
 $t$ & milliseconds & $0$ & $\inf$ \\
 $la$ & degrees & $-90$ & $+90$ \\
 $lo$ & degrees & $-180$ & $+180$ \\
 $a$ & millimeters & $0$ & $\inf$ \\
 $\phi$ & radians & $-\pi$ & $+\pi$ \\
 $\theta$ & radians & $-\pi$ & $+\pi$ \\
 $\psi$ & radians & $-\pi$ & $+\pi$ \\
 $V_x$ & m/s & $0$ & $\inf$ \\
 $V_y$ & m/s & $0$ & $\inf$ \\
 $V_z$ & m/s & $0$ & $\inf$ \\
\hline
\end{tabular}
\end{center}
\end{table}

\begin{figure}[hbt]
    \centering
    \includegraphics[width=0.40\textwidth]{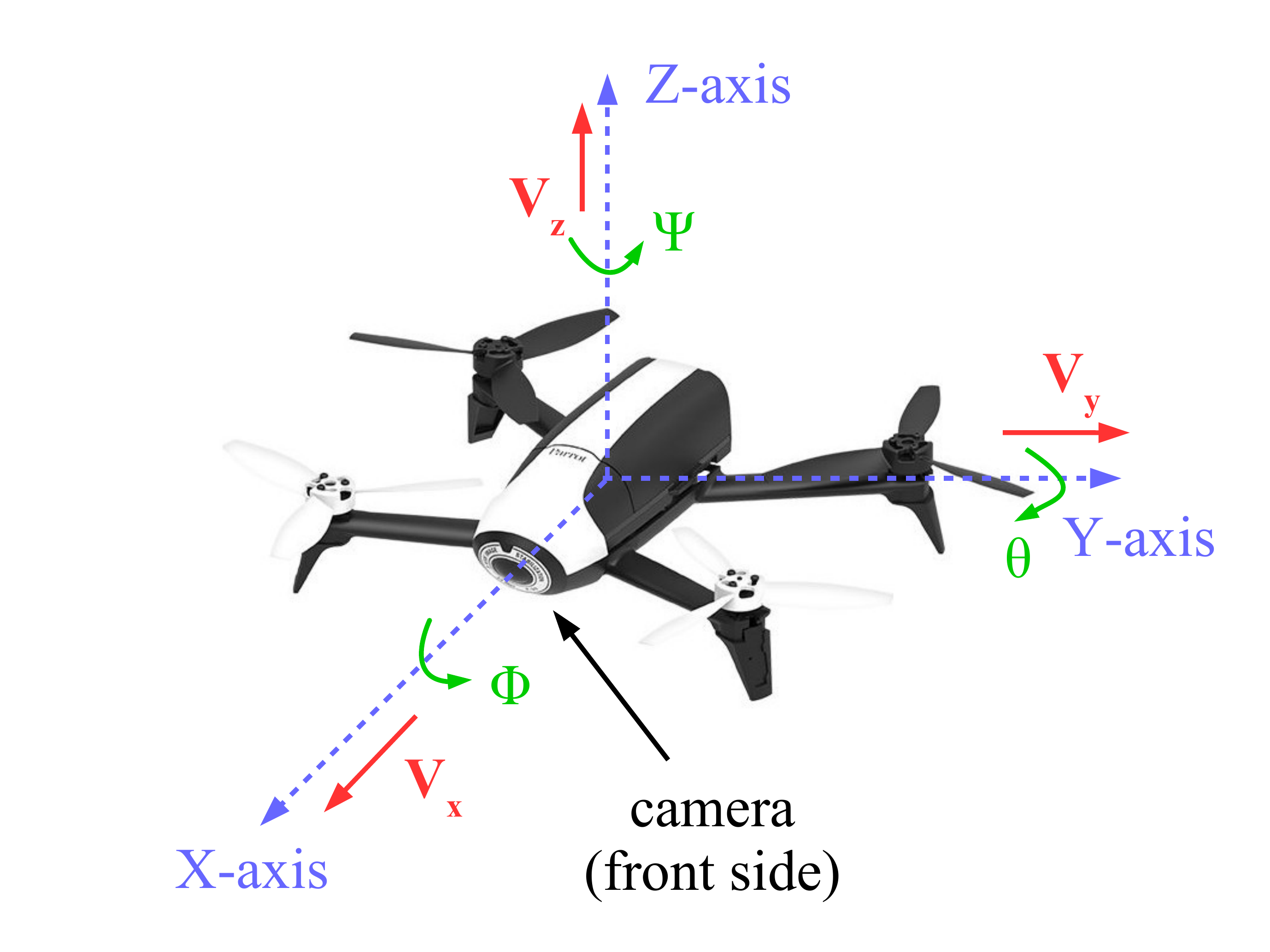}
    \caption{The body-frame of Parrot Bebop 2. [Best viewed in color]}
    \label{fig:drone}
\end{figure}

The velocities ($V_x, V_y, V_z$) and rotation angles ($\phi, \theta, \psi$) are calculated according to the UAV body-frame given in Fig. \ref{fig:drone}.

\section{EVALUATION AND ANALYSIS}
We train and evaluate mobile object detectors with our dataset. During the evaluation, we consider real-time performance rather than achieving a state-of-the-art accuracy for the sake of the applicability. Therefore, we choose two mobile object detectors (YOLOv3-Tiny \cite{redmon2018yolov3} and MobileNetv2-SSDLite \cite{sandler2018mobilenetv2}), which have a reasonable trade-off between the detection accuracy and the inference time.

\subsection{Baseline networks}
We configure YOLOv3-Tiny \cite{redmon2018yolov3} and MobileNetv2-SSDLite \cite{sandler2018mobilenetv2} for the bench-marking using the default parameters (e.g., learning rate, input size) as suggested in the original papers. We use the models that are trained on the COCO dataset as backbones.

We split the AU-AIR dataset into \%60 training, \%10 validation and \%30 testing samples. The object detectors are adapted to the total number of classes in the AU-AIR dataset (8 classes in total) by changing their last layers.

\subsection{Comparison Metrics}
To compare detection performances, we use mean average precision (mAP) that is a prominent metric in object detection \cite{lin2014microsoft, everingham2010pascal}. It is the mean of the average precision (AP) values which compute the precision score for an object category at discretized recall values over 0 to 1 \cite{everingham2010pascal}. We consider 11 different recall values as in  \cite{everingham2010pascal} and the intersection over union (IoU) threshold as 0.5.

\subsection{Results}
For benchmarking, we train YOLOv3-Tiny and MobileNetv2-SSDLite with the AU-AIR Dataset. We use the batch size of 32 and Adam optimizer with the default parameters (alpha= 0.001, beta1=0.9, beta2=0.999). The training is stopped when the validation error starts to increase. Both networks are pre-trained on the COCO dataset. In order to see the effect of the training with an aerial dataset and a natural image dataset, we also use YOLOv3-Tiny and MobileNetv2-SSDLite trained on the COCO dataset without further training with the AU-AIR dataset. The results are given in Table \ref{tbl:baselines}.

\begin{table*}[!htbp]
\caption{Category-wise average precision values of the baseline networks.
\label{tbl:baselines}}
\begin{center}
\footnotesize
\begin{tabular}{c|c|cccccccc|c}\hline
 & Training &  &  &  &  &  &   &  &  &  \\ 
Model & Dataset & Human & Car & Truck & Van & Motorbike & Bicycle  & Bus & Trailer & mAP \\  \hline \hline

YOLOV3-Tiny  & AU-AIR & 34.05 & 36.30 &  47.13 & 41.47 & 4.80 & 12.34 & 51.78 & 13.95 & 30.22 \\
MobileNetV2-SSDLite & AU-AIR & 22.86 &  19.65 & 34.74 & 25.73 & 0.01 &0.01 & 39.63 &   13.38 & 19.50 \\

YOLOV3-Tiny  & COCO & 0.01 & 0 &  0 & n/a & 0 & 0 & 0 & n/a & n/a \\
MobileNetV2-SSDLite  & COCO & 0 & 0 &  0 & n/a & 0 & 0 & 0 & n/a & n/a \\

\hline

\end{tabular}
\end{center}

\end{table*}

As shown in Table \ref{tbl:baselines}, the networks only trained on the COCO dataset have poor results. This is expected since the characteristics of natural images are significantly different from natural images. 

We observe that the AP values of motorbike and bicycle categories are significantly lower than the AP values of other categories. This fact might happen due to the class imbalance problem and the small object sizes of these categories. However, the bus category has the highest AP value, although there are fewer bus instances. This might result from the large size of bus instances in the frames. Furthermore, although the size of human instances is usually as small as the sizes of motorbike and bicycles, the AP values of the human category are relatively higher than these classes. This fact might be a consequence of the high number of human instances. There is no available AP values for the van and trailer categories in Table \ref{tbl:baselines} since they do not exist in the COCO dataset.

\begin{figure}[!t]
    \centering
    \includegraphics[width=0.49\textwidth]{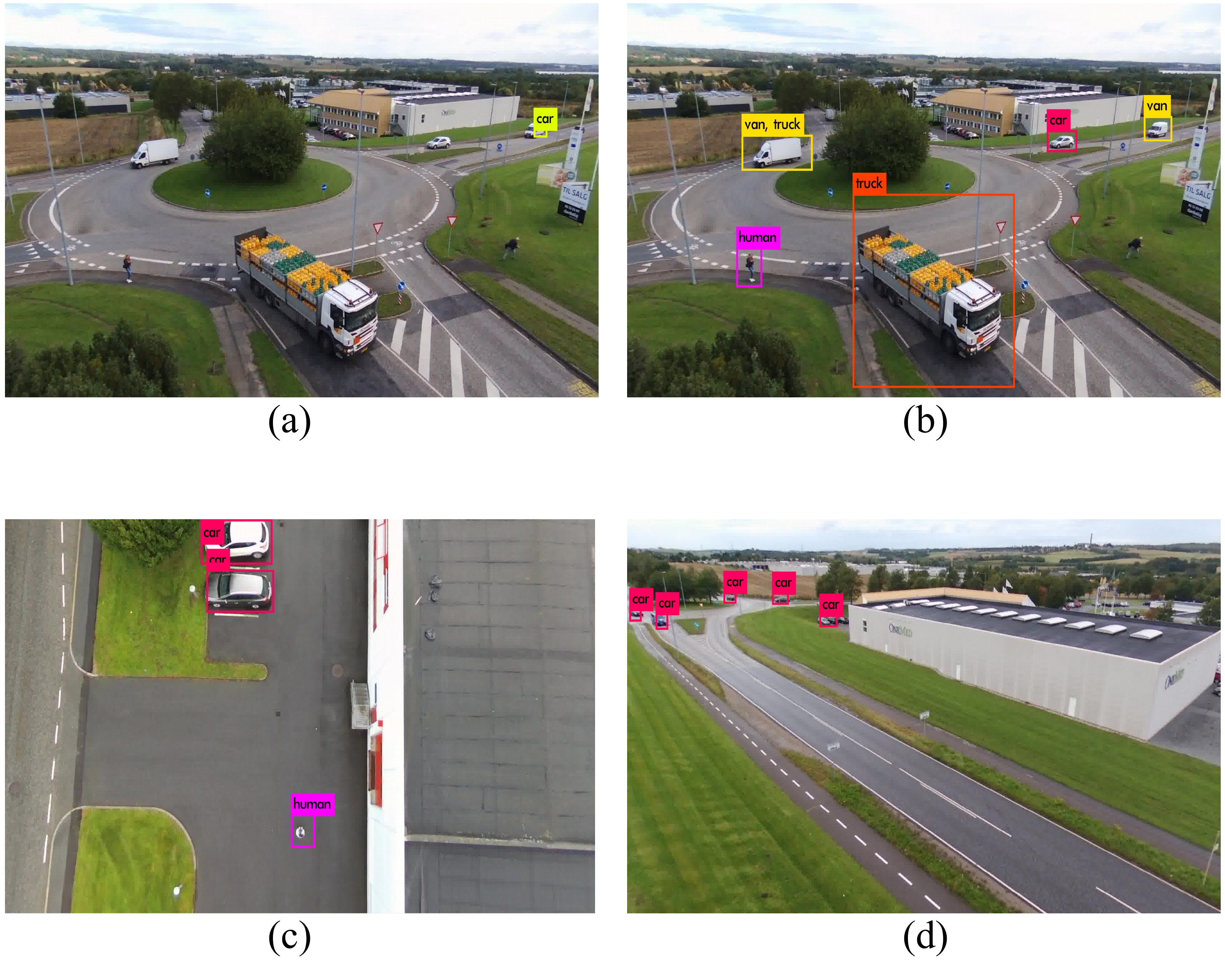}
    \caption{Sample results of YOLOv3-Tiny networks. The network trained on only the COCO dataset has poor detection performance (a). The network trained on the AU-AIR dataset with the COCO pre-training has better results for the same input (b). It can also detect objects in complete bird-view images, which appears significantly different from natural images. The network trained on AU-AIR dataset can also find objects which suffer from the perspective (d). [Best viewed in color]}
    \label{fig:sample_results}
\end{figure}

The baselines trained on the AU-AIR dataset are good at finding objects in aerial images that are captured at different altitudes and view angles. Qualitative results can be seen in Fig. \ref{fig:sample_results}.

Among the baselines, YOLOv3-Tiny has higher AP values and mAP value compared to MobileNetv2-SSDLite. There is no significant difference between inference times (17.5 FPS and 17 FPS for YOLOv3-Tiny and MobileNetv2-SSDLite on TX2, respectively).

\section{DISCUSSION}

Since the number of instances of each object category is imbalanced in the AU-AIR dataset (Fig. \ref{fig:classes_vs_numsamples}), we consider several methods to solve the imbalanced class problem in the next version of the dataset. As a first step, we will try to collect more data to balance the number of instances. Besides, we may consider adding synthetic data (i.e., changing the brightness of images, translation, rotation) to increase the number of object categories which has a low number of samples in the current version.

We use AMT to annotate objects in images. Although three different people annotate one image and the annotations are manually checked by ourselves, there might be still overlooked samples that have weak annotations (e.g., unlabelled instances, loose bounding box drawings). Therefore, we consider using a three-step workflow proposed by Su et al. \cite{su2012crowdsourcing}. In this workflow, the first worker draws a bounding box around an instance, the second worker verifies whether the bounding box is correctly drawn, and the third worker checks whether all object instances are annotated.

Unlike other UAV object detection datasets, ours includes sensor data corresponding to each frame. In this work, we give a baseline only for object annotations and visual data. As future work, more baselines may be added to encourage research using sensor data (e.g., navigation and control of a UAV, object detection using multi-modal data). Also, we can add more visual sensors, such as multi-spectral cameras.

We have used a ready-to-fly quadrotor (i.e., Parrot Bebop 2) to collect the whole dataset. We also consider collecting more samples from other platforms (e.g., different types of UAVs) using cameras that have different resolutions and frame rates.

In this dataset, traffic surveillance is the primary context. In future work, we consider increasing the number of environment contexts to increase diversity in the dataset.


\section{CONCLUSIONS}
In this work, we propose the AU-AIR dataset that is a multi-modal UAV dataset collected in an outdoor environment. Our aim is to fill the gap between computer vision and robotics having a diverse range of recorded data types for UAVs.  Including visual data, object annotations, and flight data, it can be used for different research fields focused on data fusion. 

We have emphasized the differences between natural images and aerial images affecting the object detection task. Moreover, since we consider real-time performance and applicability in real-world scenarios, we have created a baseline, including two mobile object detectors in the literature (i.e., YOLOv3-Tiny \cite{redmon2018yolov3} and MobileNetv2-SSDLite \cite{sandler2018mobilenetv2}). In our experiments, we showed that mobile networks trained on natural images have trouble in detecting objects in aerial images.

\addtolength{\textheight}{-12cm}   





\bibliographystyle{IEEEtran}
\bibliography{references}

\end{document}